\pdfoutput=1

\documentclass[11pt]{article}

\usepackage{emnlp2021}

\usepackage{times}
\usepackage{latexsym}

\usepackage[T1]{fontenc}

\usepackage[utf8]{inputenc}

\usepackage{microtype}

\usepackage{booktabs}
\usepackage{graphicx}
\usepackage{fontawesome}
\usepackage{amsmath}
\usepackage{multirow}

\title{Revisiting Robust Neural Machine Translation:\\A Transformer Case Study}

\author{Peyman Passban$^{1, }$\Thanks{ Equal contribution.}\hspace{1.6mm}$^,$\hspace{0.2mm}\Thanks{ Work done while Peyman Passban was at Huawei.} \hspace{2mm} Puneeth S.M. Saladi$^{1,2,}$\footnotemark[1] \hspace{2mm} Qun Liu$^1$ \\\\
    $^1$Huawei Noah's Ark Lab \\
    $^2$Faculty of Mathematics, University of Waterloo \\
    \texttt{passban.peyman@gmail.com} \\
    \texttt{\{puneeth.saladi,qun.liu\}@huawei.com}}

\begin{document}
\maketitle
\begin{abstract}
Transformers \cite{transformer} have brought a remarkable improvement in the performance of neural machine translation (NMT) systems but they could be surprisingly vulnerable to noise. In this work, we try to investigate how noise breaks Transformers and if there exist solutions to deal with such issues. There is a large body of work in the NMT literature on analyzing the behavior of conventional models for the problem of noise but Transformers are relatively understudied in this context. Motivated by this, we introduce a novel data-driven technique called Target Augmented Fine-tuning (TAFT) to incorporate noise during training. This idea is comparable to the well-known fine-tuning strategy. Moreover, we propose two other novel extensions to the original Transformer: Controlled Denoising (CD) and Dual-Channel Decoding (DCD), that modify the neural architecture as well as the training process to handle noise. One important characteristic of our techniques is that they only impact the training phase and do not impose any overhead at inference time. We evaluated our techniques to translate the English--German pair in both directions and observed that our models have a higher tolerance to noise. More specifically, they perform with no deterioration where up to $10$\% of \textit{entire} test words are infected by noise.
\end{abstract}

\section{Introduction}
NMT is the task of transforming a source sequence into a new form in a particular target language using deep neural networks. Such networks commonly have an encoder-decoder architecture \cite{cho-etal-2014-properties,cho-etal-2014-learning,sutskever2014sequence}, in which an encoder maps a given input sequence to an intermediate representation and a decoder then uses this representation to generate candidate translations. Both encoder and decoder are neural networks that are trained jointly. Due to the sequential nature of the NMT task, early models usually relied on recurrent architectures \citep{yang2020survey}, or benefited from the sliding feature of convolutional kernels to encode/decode variable-length sequences \citep{kalchbrenner2014convolutional,gehring-etal-2017-convolutional}. 

Recently, Transformers \citep{transformer} have shown promising results for NMT and become the new standard in the field. They follow the same concept of encoding and decoding but in a relatively different fashion. A Transformer is fundamentally a feed-forward model with its unique neural components (self-attention, layer norm, etc) that altered the traditional translation pipeline accordingly. It is expected if such a new architecture would behave differently than its recurrent or convolutional counterparts, and our goal in this research is to study this aspect in the presence of noise.   

NMT engines trained on clean samples provide high-quality results when tested on similarly clean texts, but they break easily if noise appears in the input \citep{michel-neubig-2018-mtnt}. They are not designed to handle noise by default and Transformers are no exception. Many previous works have focused on this issue and studied different architectures \citep{li2019findings}. However, in this work, we particularly focus on Transformers\footnote{We assume that the reader is already familiar with the Transformer architecture.} as they are relatively new and to some extent understudied. 

A common approach to make NMT models immune to noise is \textit{fine-tuning} (\textit{FT}), where a noisy version of input tokens is intentionally introduced during training and the decoder is forced to generate correct translations despite deformed inputs. \textit{FT} is quite useful for almost all situations but it needs to be run with an optimal setting to be effective. In our experiments, we propose a slightly different learning-rate scheduler to improve \textit{FT}. We also define a new extension that not only modifies input words but also adds complementary tokens to the target side. We refer to this extension as \textit{Target Augmented Fine-Tuning} (\textit{TAFT}), which is the first contribution of this paper. 

In our study, we realized that data augmentation techniques (\textit{FT} and \textit{TAFT}) might not be sufficient enough and we need a compatible training process and neural architecture to deal with noise. Therefore, we propose \textit{Controlled Denoising} (\textit{CD}) whereby noise is added to source sequences during training and the encoder is supposed to fix noisy words before feeding the decoder. This approach is implemented via an auxiliary loss function and is similar to adversarial training. \textit{CD} is our second contribution. 

\textit{CD} only takes care of noise on the encoder side. We also propose a \textit{Dual-Channel Decoding} (\textit{DCD}) strategy to study what happens if the decoder is also informed about the input noise. \textit{DCD} supports multi-tasking through a $2$-channel decoder that samples target tokens and corrects noisy input words simultaneously. This form of fusing translation knowledge with noise-related information has led to interesting results in our experiments. \textit{DCD} is the third and last contribution of this work. 

The remainder of the paper is organised as follows: We first review previously reported solutions for the problem of noise in NMT in Section \ref{back}. Then, we present details of our methods and the intuition behind them in Section \ref{method}. To validate our methods, we report experimental results in Section \ref{exp}. Finally, we conclude the paper and discuss possible future directions in Section \ref{cfw}.

\section{Related Work}\label{back}
Fine-tuning (\textit{FT}) is one of the most straightforward and reliable techniques to protect NMT systems from noise. \citet{berard2019naver}, \citet{dabre-sumita-2019-nicts}, and \citet{helcl2019cuni} studied its impact and showed how it needs to be utilized to boost NMT quality.

Adversarial training is another common solution to build noise-robust models. \citet{doubly-adv} proposed a gradient-based method to construct adversarial examples for both source and target samples. Source-side inputs are supposed to attack the model while adversarial target inputs help defend the translation model. In their model, a candidate word is replaced with its semantically-close peer to introduce noise. This way, the neural engine visits different forms of the same sample, which extends its generalization. In other words, the network is trained to deliver the same, consistent functionality even though it is fed with different forms of a sample. Although this strategy showed promising results, in our setting we replace input words with \textit{real} noisy candidates instead of synonyms or semantically-related peers. We find this way of adding noise more realistic and closer to real-world scenarios. 

\citet{karpukhin-etal-2019-training} experimented another idea by generating adversarial examples using synthetic noise. Their proposed architecture relies on Transformers but the encoder is equipped with a character-based convolutional model \citep{kim2015character}. This work is one of the few attempts that studied Transformers' behaviour in the presence of noise. However, their results are based on relatively small datasets. We know that NMT models' performance could change proportionally with a change in the size of training sets. Therefore, we used larger datasets in our experiments. 

The application of adversarial training is not limited to the aforementioned examples. \citet{towards-robust-nmt} defined additional loss functions which force the encoder and decoder to ignore perturbations and generate clean outputs. This idea is similar to our \textit{CD} approach, but the underlying architecture is different. \citet{towards-robust-nmt} only reported results on recurrent NMT models. 

Providing better representations is as important as designing tailored training strategies for noise-robust models. A group of researchers focused on how different segmentation schemes and encoding techniques can play a role. \citet{sennrich2015neural} and  \citet{michel-neubig-2018-mtnt} showed that subwords are better alternatives than surface forms (words) to handle perturbations and out-of-vocabulary words. \citet{belinkov2018synthetic} comprehensively studied this by using different character- and subword-based representations in different architectures. \citet{sakaguchi17robsut} also carried out a similar investigation where they proposed a new encoding that is invariant to the order of characters.  

Besides these approaches, translating noisy inputs can be viewed as a two-pass process performed via two connected neural networks. The first one acts as a monolingual translator to correct noisy inputs and the second one is an engine that consumes denoised sequences to generate clean translations \citep{text-denoising-mlm,multi-task}. This idea can be implemented as an end-to-end, differentiable solution or as a pipeline, but it should be noted that such a mechanism could be hard to deploy or slow(er) to run in practice.

\section{Methodology}\label{method}
This section covers details of our proposed methods. \textit{FT} is a well-known technique so we skip its details and only focus on \textit{TAFT}, our own extension of it (Section \ref{ft}). Besides \textit{FT} and \textit{TAFT} that leverage data, we introduce \textit{CD} (Section \ref{cd}) and \textit{DCD} (Section \ref{dcd}), which modify the training procedure as well as the neural architecture of Transformers.  

\subsection{Fine-Tuning Transformers}\label{ft}
\textit{FT} simply exposes an already-trained translation engine to noise during training in the hope of extending its coverage at test time. This simple idea is quite effective, but it requires to be run with an optimal setting, e.g. the type/amount of noise added to the training set directly impacts performance. It is also crucial to find an optimal number of iterations. Overrunning \textit{FT} could hurt quality and be as costly as training a new model, and running it for an insignificant number of iterations might not be enough to reveal its power. Clearly, a better choice of these hyper-parameters leads to better results, but in addition to this empirical side of \textit{FT}, we realized that it can be boosted even more via a simple modification.

\textit{FT} only alters source sequences. In our extension (\textit{TAFT}), we change the target side as well by appending clean versions of perturbed source words to the target sequence. Table \ref{t:1} provides an example for this form of data augmentation. An ordinary model works with clean forms of source and target tokens, as shown in the first block. The second source word `\textbf{anderen}' is randomly selected to be substituted with its noisy version `\textcolor{red}{\textit{andare}}. In \textit{FT}, a source sequence including this noisy form (or its preprocessed version) is sent to the translation engine but the target sequence remains untouched. \textit{TAFT} works with a slightly different data format where the source sequence includes the noisy input and at the same time its clean version (namely `\textcolor{black}{\textbf{anderen}}') is appended to the original target sequence. With this simple technique, the NMT model is forced to generate translations, spot noisy source tokens, and correct them all together. This could be considered as an implicit form of multi-tasking without changing the neural architecture. The fusion of translation and correction knowledge on the decoder side seems to be useful (see our experimental results in Tables \ref{t:w2b} and \ref{t:b2b}).
\begin{table}[t]
\small
\begin{tabular}{l}
\toprule
\multicolumn{1}{c}{\textit{Original}} \\ \hline
alle \textbf{anderen} waren anderer meinung . \\ 
all of the others were of a different opinion . \\ \hline
\multicolumn{1}{c}{\textit{FT}} \\ \hline
alle \textcolor{red}{and@@ are} waren anderer meinung . \\ 
all of the others were of a different opinion . \\ \hline
\multicolumn{1}{c}{\textit{TAFT}} \\ \hline
alle \textcolor{red}{and@@ are} waren anderer meinung . \\ 
all of the others were of a different opinion . \textcolor{black}{\textbf{anderen}}\\ 
\bottomrule
\end{tabular}
\caption{How \textit{FT} and \textit{TAFT} process training examples. Noise is added to the boldfaced word. This example is selected from our German$\rightarrow$English corpus. In each block, the first sequence is from the source and the second sequence is from the target side. Sequences are pre-processed and tokenized.}
\label{t:1}
\end{table}

It should be noted that at test time in \textit{TAFT}, the engine only generates tokens of the target sequence, i.e. it stops decoding as soon as it visits the end of the target sequence. Generating target tokens together with clean source words is a training-time technique to improve the robustness of the model. Therefore, this extension does not slow down the model or change anything about it at inference. Moreover, if any segmentation scheme such as byte-pair encoding (\textit{bpe}) \cite{sennrich2015neural} is applied to input words during preprocessing, the noisy form also needs to be preprocessed accordingly. The same rule applies to the clean form appended to the target sequence too, namely it needs to follow the segmentation scheme of the target side. In Table \ref{t:1}, the noisy form `\textcolor{red}{\textit{andare}}' is segmented into `\textcolor{red}{\textit{and@@}}' and `\textcolor{red}{\textit{are}}' via the source-side \textit{bpe} model and the correct form `\textcolor{black}{\textbf{anderen}}' is appended as is because the target \textit{bpe} model was able to recognize it as an existing entry. 

\subsection{Controlled Denoising}\label{cd}
\textit{FT} and \textit{TAFT} have no control over the encoder's output, and it is assumed that the decoder alone is powerful enough to handle representations of noisy inputs and deliver correct translations. This assumption might fail in practice, so we place a filter after the encoder to purify source representations before sending them to the decoder. We refer to this process as \textit{Controlled Denoising} (\textit{CD}).   

The idea behind \textit{CD} is to force the encoder to correct its noisy representations. To implement this mechanism, we accompany the main encoder (the one that is connected to the decoder) with an auxiliary, pre-trained encoder. These two encoders have an identical architecture and work with the same vocabulary. The main encoder consumes sequences that may include noisy tokens but the auxiliary one is always fed with clean sequences. These two encoders meet before the first layer of the decoder to ensure they both generate the same representations regardless of any discrepancies that may occur in their input. If there appears a noisy token in the input of the main encoder its output would differ from that of the auxiliary one. Therefore, we match the output of these two encoders via a loss function to ensure the main encoder is able to handle/ignore the input noise.

\begin{figure}[ht]
    \centering
    \includegraphics[width=\linewidth]{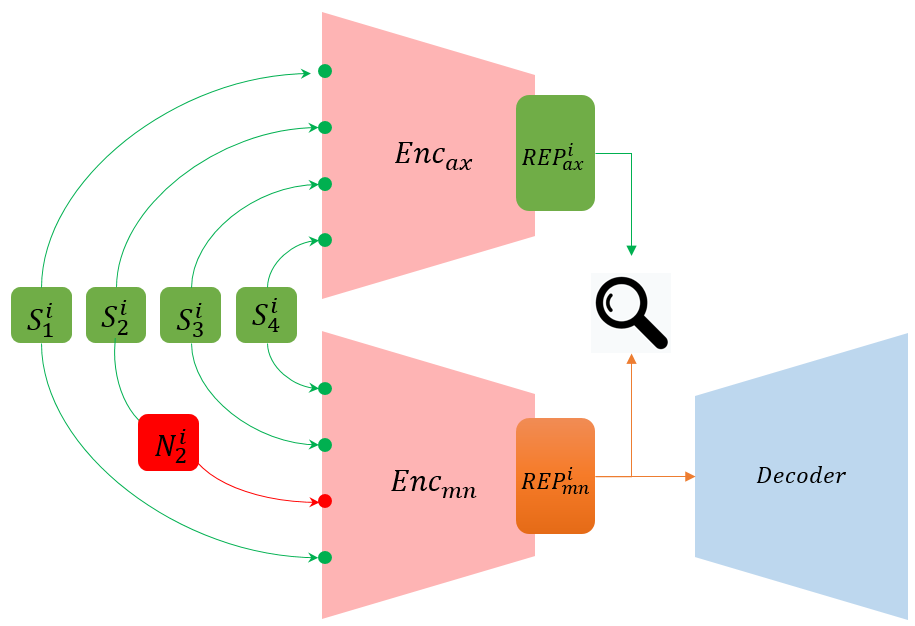}
    \caption{The \textit{CD} architecture. $S^i_j$ is the $j$-th token of the $i$-th sequence whose perturbed form is $N^i_j$. $\textrm{{\fontfamily{pcr}\selectfont REP}}_{mn}$ and $\textrm{{\fontfamily{pcr}\selectfont REP}}_{ax}$ show the representations of the input sequences generated by the \textit{main} and \textit{auxiliary} encoders, respectively. These two representations are compared to each other via a loss function (denoted with {\footnotesize  \faSearch}) to ensure that the main encoder is able to handle noisy inputs.}
    \label{fig:cd}
\end{figure}

In the absence of noise, the main encoder mimics the auxiliary encoder, but when noise is added  the main encoder's outputs may deviate from expected ones. The loss function in between the two encoders helps the main encoder correct itself and push its outputs closer to clean representations (auxiliary encoder's outputs). This architecture is illustrated in Figure \ref{fig:cd}. 

Conventional recurrent and convolutional encoders usually squeeze the representation of the input sequence into a single vector, but Transformers due to their non-recurrent architecture perform differently and instead generate $s$ vectors if the input sequence consists of $s$ tokens. This makes the comparison between outputs of the main and auxiliary encoders challenging. Because, when an input sequence is perturbed with noise the length of the noisy sequence could vary from the clean one, e.g. one token can be added/dropped or the noisy token can be decomposed into multiple units via \textit{bpe} (as shown in Table \ref{t:1}). In such cases, the shape of encoders' outputs does not even match and a vector-to-vector comparison is impossible. To handle these issues, we learn a dedicated representation for \textit{the entire input sequence}, so comparing outputs would be straightforward. We do this form of sequence modelling by following the same idea proposed for the {\fontfamily{pcr}\selectfont CLS} token in \citet{Devlin2019BERTPO}. We refer to this sequence-level representation as {\fontfamily{pcr}\selectfont REP} in our setting. 

In Figure \ref{fig:cd}, the inputs to the main and auxiliary encoders are [$S^i_1, N^i_2, S^i_3, S^i_4$] and [$S^i_1, S^i_2, S^i_3, S^i_4$], respectively, and their sequence-level representations are $\textrm{{\fontfamily{pcr}\selectfont REP}}_{mn}$ and $\textrm{{\fontfamily{pcr}\selectfont REP}}_{ax}$. If our Transformer encoder is fed with $s$ tokens it returns $s+1$ vectors with the last one being {\fontfamily{pcr}\selectfont REP}. This token is only used for comparison purposes between encoders and is not sent to the decoder. 

To train our models with \textit{CD}, we extend the original translation loss ($\mathcal{L}_{tr}$) with an additional term, $\mathcal{L}_{CD}$, as defined in Equation \ref{eq:cdloss}: 
\begin{equation}\label{eq:cdloss}
    \begin{split}
    \mathcal{L} =& \hspace{1mm} \alpha \mathcal{L}_{tr} + \beta \mathcal{L}_{CD} 
    \\
    \mathcal{L}_{_{CD}} =& \sum_{i} \textrm{MSE}( \textrm{{\fontfamily{pcr}\selectfont REP}}_{mn}^i,\textrm{{\fontfamily{pcr}\selectfont REP}}_{ax}^i)
    \end{split}
\end{equation}
where MSE() is the mean-square error and $\textrm{{\fontfamily{pcr}\selectfont REP}}_{mn}^i$ and  $\textrm{{\fontfamily{pcr}\selectfont REP}}_{ax}^i$ are the sequence-level representations of the $i$-th training sequence generated by our two encoders. $\alpha$ and $\beta$ are weights to adjust the contribution of each loss function in the training process. Check Section \ref{exp} for their values. 

As previously mentioned, the auxiliary encoder is a pre-trained model (trained on the clean/original sentences of the same dataset with an identical vocabulary set and architecture as the main encoder) and is only used to generate reference representations for the main encoder, so its parameters are not updated during training and only main encoder and decoder's parameters are impacted in the back-propagation phase. It should also be noted that the auxiliary encoder is used during training and is decoupled later for inference. Therefore, the size of the final model and inference time remain the same as in the original Transformer. 

\subsection{Dual-Channel Decoding}\label{dcd}

\begin{figure}[ht]
    \centering
    \includegraphics[width=0.49\textwidth]{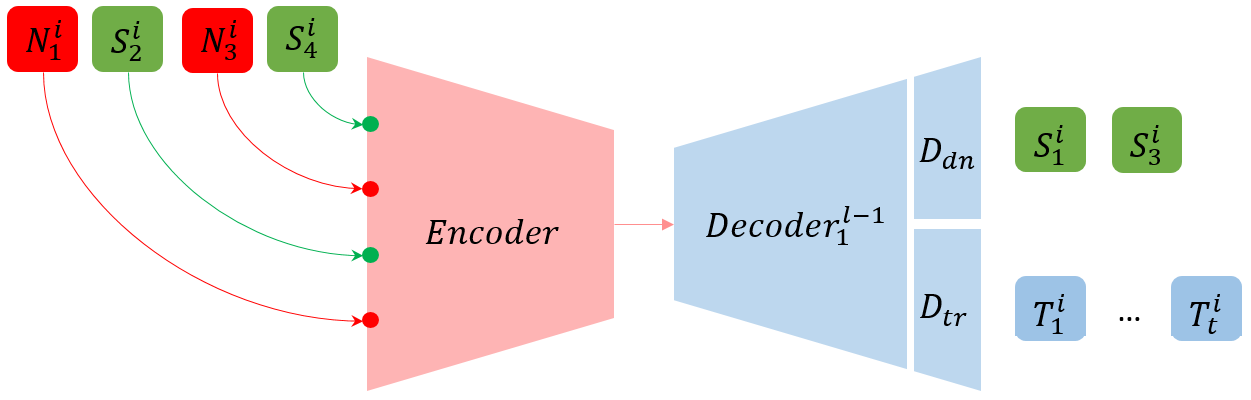}
    \caption{The DCD architecture. $D_{dn}$ generates [$S^i_1, S^i_3$] whcih are the clean version of noisy tokens on the input side and $D_{tr}$ samples $t$ tokens from the target vocabulary to generate the final translation [$T^i_1,...,T^i_t$]. The decoder has $l$ layers where $(l-1)$ of them are shared.}
    \label{fig:DCD}
\end{figure}

This approach relies on the idea of data augmentation and multi-tasking, and tries to break down noise-robust NMT into two tasks of \textit{denoising} and \textit{translation}. Our \textit{DCD} architecture has two decoder channels: One for generating the rectified noisy tokens ($D_{dn}$) detected in the input sequence and the other one ($D_{tr}$) for actual translations. The extra task defined in addition to translation is meant to guide the decoder by providing richer information and make it robust against input noise. This architecture is illustrated in Figure \ref{fig:DCD}. 

The original Transformer decoder has $l$ layers. In our \textit{DCD} extension, the first $l-1$ layers are shared in between two tasks, but the last layer has dedicated components for each. $D_{dn}$ is a unique decoder layer that is responsible to generate clean forms of any noisy token that can appear in the input, e.g. in Figure \ref{fig:DCD} the first ($N^i_1$) and third ($N^i_3$) tokens are perturbed so $D_{dn}$ generates $S^i_1$ and $S^i_3$ (their clean versions) as its output. $D_{dn}$ is trained via a dedicated loss function ($\mathcal{L}_{dn}$) designed for this task. On the other side, $D_{tr}$ is another decoder layer that shares no parameter with $D_{dn}$. This layer is placed over the decoder's lexicon and samples target words to generate the final translation. This layer is connected to $\mathcal{L}_{tr}$ to penalize incorrect translations. In this setting, the final loss function is a composition of two losses, as defined in Equation \ref{eq:DCD}: 
\begin{equation}\label{eq:DCD}
    \mathcal{L} = \hspace{1mm} \alpha \mathcal{L}_{tr} + \beta \mathcal{L}_{dn}
\end{equation}

The main purpose of having such a semi-shared architecture for each task is to benefit from the power of multi-tasking. Both $D_{tr}$ and $D_{dn}$ are triggered with a mixture of information about translation and denoising provided by the first $l-1$ layers of the decoder; then they use their dedicated modules/parameters to generate different outputs for their particular task. Similar to \textit{CD}, this technique is also employed during training and at inference we do not require $D_{dn}$. This should ensure similar memory consumption and inference speed as the vanilla Transformer model.

\section{Experimental Study}\label{exp}
\textbf{Datasets}\hspace{4mm} To evaluate our models, we trained engines to translate the English--German (En--De) pair in both directions. In the interest of fair comparisons, we used the same datasets as the original Transformer \citep{transformer}, so our training set is the {\fontfamily{pcr}\selectfont WMT-14} dataset\footnote{\url{http://statmt.org/wmt14/translation-task.html}} with $4.5$M parallel sentences and for development and test sets we used {\fontfamily{pcr}\selectfont newstest-13} and {\fontfamily{pcr}\selectfont newstest-14}, respectively. 

Nowadays, almost all state-of-the-art NMT models rely on subwords. We also followed the same tradition and preprocessed the target side of our datasets with \textit{bpe} \citep{sennrich2015neural}. For the source side, we used different segmentation schemes with different granularities as we are studying the impact of noisy inputs. Our source tokens can appear in surface forms (words) or can be segmented into \textit{bpe} tokens or oven characters. We refer to these settings as \textit{word2bpe}, \textit{bpe2bpe}, and \textit{char2bpe}, respectively. 

The size of the lexicon generated by our \textit{bpe} model also matches the setting proposed for the original Transformer model \citep{transformer}. For the \textit{word2bpe} setting, we keep the top $48$K frequent words for each English and German sides and ignore the rest by substituting with a special {\fontfamily{pcr}\selectfont UNK} token. This configuration is learned through an empirical study to maximize translation quality.\\\\
\textbf{Hyper-parameters}\hspace{4mm} We carried out multiple experiments to study how each of \textit{word2bpe}, \textit{bpe2bpe}, and \textit{char2bpe} settings react in conjunction with our models and what values should be used for hyper-parameters. 

In these experiments, we did not change the neural architecture for the \textit{FT} and \textit{TAFT} models and only trained translation engines with augmented datasets. We realized that fine-tuning can be improved if we slightly change the learning-rate scheduling. The original Transformer uses the Noam scheduling \cite{transformer} that employs a linear warm-up strategy followed by a decaying function. We changed it to a simple exponential staircase decay with an initial learning rate of $0.001$ and a decay rate of $0.5$ after every $5,000$ steps. 

\begin{figure}[ht]
    \centering
    \includegraphics[width=0.9\linewidth]{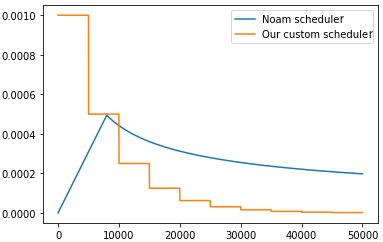}
    \caption{Comparing Noam to our custom learning rate scheduler. The \textit{y} axis represents the learning rate and the \textit{x} axis shows the number of steps.}
    \label{fig:lr}
\end{figure}

As our translation engines are already trained and we only need to fine-tune them, we can ignore the warm-up strategy. We can start from a non-zero value and carefully decrease the learning rate until models converge. We fine-tune all our models for $50,000$ steps with this scheduler. Figure \ref{fig:lr} illustrates the difference between Noam and our scheduler. 
 
Apart from \textit{char2bpe} that uses the batch size of $12,288$, all other settings process $4,096$ tokens in each batch. \textit{char2bpe} is a character-based model and the more characters it processes, the better performance it gains. We set $\alpha$ and $\beta$ (loss weights) to [$0.75$, $0.25$] and [$0.9$, $0.1$] for \textit{CD} and \textit{DCD}, respectively. Our models are trained using Adam \citep{kingma2014adam}. If the value of any other hyper-parameter (such as the embedding dimension etc) is not clearly mentioned in the paper that means we use the original value of it proposed by \citet{transformer}.
\subsection{Introducing Noise}
To train/test our engines, we need to perturb source sentences by injecting noise. A noisy word can be created by adding, dropping, or replacing a character in a word or by imposing any other deformations \citep{towards-robust-nmt,michel-neubig-2018-mtnt}. However, all these techniques artificially produce new forms that might not necessarily reflect real-world noise. We thus use a particular type of noise which is known as \textit{natural noise} in the literature \citep{belinkov2018synthetic}. This form is an error that can naturally appear in any text. Researchers collected lists\footnote{\url{https://github.com/ybisk/charNMT-noise/tree/master/noise}} of frequently-occurring mistakes/typos in different languages from existing corpora and made them available. In our experiments, we randomly pick a candidate word and retrieve its noisy version from the aforementioned lists. This way we could ensure that our noisy dataset is representative of what we may encounter in real life.

To create our training sets, we randomly select $50$\% of \textit{sentences} to perturb with noise. We only destroy one word in each sentence. Noise is added to surface forms, so if the neural encoder is designed to work with a different granularity, all necessary preprocessing steps are applied accordingly, e.g. in Table \ref{t:1}, first the candidate word (\textbf{anderen}) is perturbed (\textit{andare}), then \textit{bpe} is applied to the noisy form to have a consistent input (\textcolor{red}{\textit{and@@ are}}) with the encoder's vocabulary.  

To add noise to our test sets we have a slightly different and relatively aggressive approach. We are interested in challenging our models to see if they can tolerate high volumes of noise, so we created $4$ noisy test sets in which $5$\%, $10$\%, $20$\%, and $30$\% of entire \textit{words} (not sentences) are destroyed. Adding noise based on the percentage of words instead of sentences makes translation quite challenging because perturbing for example $10$\% of sentences (with one noisy word) in our $3003$-sentence test set only generates $300$ noisy words whereas this number would be around $7000$ if we perturb $10$\% of the entire words. Unlike the training setting where we only perturb one word in a sentence, in the test setting, multiple words can be impacted. 

Since, this is the first work (to the best of our knowledge) that particularly studies Transformers  for their ability to tackled noise we only selected to work with \textit{natural} noise, which seems to be the most realistic form. However, our work can be  extended by investigating the impact of other famouse noise classes such as \textit{Swap}, \textit{Mid}, \textit{Rand} etc. For detailed information on noise classes see \citet{belinkov2018synthetic}, \citet{impact-of-noise-types}, and \citet{michel-neubig-2018-mtnt}.

\subsection{Baseline Models} 
\begin{table*}[ht]
\centering
\begin{tabular}{l l c c c c c }
\toprule
\textbf{} & \textbf{} & \textbf{0\%} & \textbf{5\%} & \textbf{10\%} & \textbf{20\%} & \textbf{30\%} \\\hline

\multirow{4}{2em}{En$\rightarrow$De} & \textit{word2bpe} & $28.48$ & $22.21$ (\textcolor{red}{-$22$\%}) & $17.05$ (\textcolor{red}{-$40$\%}) & $10.28$ (\textcolor{red}{-$64$\%}) & \hspace{1.4mm}$5.99$ (\textcolor{red}{-$79$\%})\\ 

 & \textit{bpe2bpe} & $28.46$ & $24.82$ (\textcolor{red}{-$13$\%}) & $21.58$ (\textcolor{red}{-$24$\%}) & $15.98$ (\textcolor{red}{-$44$\%}) & $11.89$ (\textcolor{red}{-$58$\%})\\
 
 & \textit{char2bpe} & $26.07$ & $24.23$ (\textcolor{red}{-$7$\%}) & $21.84$ (\textcolor{red}{-$16$\%}) & $18.37$ (\textcolor{red}{-$30$\%}) & $15.01$ (\textcolor{red}{-$42$\%})\\
 
 & \textit{ConvT} & $25.46$ & $22.55$ (\textcolor{red}{-$11$\%}) & $20.13$ (\textcolor{red}{-$21$\%}) & $14.9$ (\textcolor{red}{-$41$\%}) & $11.29$ (\textcolor{red}{-$56$\%})\\
 \hline
 \multirow{5}{3em}{De$\rightarrow$En} & \textit{word2bpe} & $25.94$ & $23.28$ (\textcolor{red}{-$10$\%}) & $20.32$ (\textcolor{red}{-$22$\%}) & $15.79$ (\textcolor{red}{-$39$\%}) & $12.00$ (\textcolor{red}{-$54$\%})\\
 
 & \textit{bpe2bpe} & $28.04$ & $24.87$ (\textcolor{red}{-$11$\%}) & $21.61$ (\textcolor{red}{-$23$\%}) & $16.11$ (\textcolor{red}{-$43$\%}) & $11.48$ (\textcolor{red}{-$59$\%})\\
 
 & \textit{char2bpe} & $26.59$ & $25.01$ (\textcolor{red}{-$6$\%}) & $22.73$ (\textcolor{red}{-$15$\%}) & $19.42$ (\textcolor{red}{-$27$\%}) & $15.93$ (\textcolor{red}{-$40$\%})\\
 
 & \textit{ConvT} & $27.08$ & $24.01$ (\textcolor{red}{-$11$\%}) & $21.39$ (\textcolor{red}{-$21$\%}) & $16.44$ (\textcolor{red}{-$39$\%}) & $11.59$ (\textcolor{red}{-$57$\%})\\
\bottomrule
\end{tabular}
\caption{BLEU scores of baseline models. Numbers inside parentheses show how much noise impacts models' performance. The first row shows the percentage of perturbed test words. \textit{All} our encoders and decoders have $8$ and $4$ layers, respectively. The \textit{char2bpe} models consumes one character at a time. The \textit{ConvT} model consumes one word at each step but applies a convolutional operation over all characters of the word before feeding it to the first encoder layer.}
\label{t:baseline}
\end{table*}
As our baseline, we trained a Transformer with a slight modification in its architecture. \citet{kasai2020deep} conducted research and showed that the number of encoder and decoder layers do not necessarily need to match and we can have imbalanced Transformers with deep(er) encoders and shallow(er) decoders. Inspired by that work, we increased the number of encoder layers\footnote{The original Transformer architecture \citep{transformer} proposes $6$ encoder and $6$ decoder layers.} from $6$ to $8$ and decreased the number of   decoder layers from $6$ to $4$. Our Transformer still has $12$ layers in total, but the encoder is more powerful which is favourable in our scenario. Noise appears on the source side and we require better encoders to tackle this. Based on our experiments, the $8$--$4$ configuration ($\mathcal{T}^8_4$) is able to handle noise better than the $6$--$6$ version and all other variants. $\mathcal{T}^8_4$ is our baseline for all experiments and our other novel architectures are also implemented based on the $8$--$4$ setting.  

\citet{belinkov2018synthetic} and \citet{karpukhin-etal-2019-training} used a convolutional, character-based encoder and showed that this improves the robustness of NMT models. They tested this configuration with relatively small datasets or recurrent architectures. We adapted the same idea and equipped the Transformer model with the same convolutional module. This model, which is referred to as \textit{ConvT} in this paper, is another baseline for our experiments. Similar to \citet{karpukhin-etal-2019-training}, character embeddings have $256$ dimensions and the convolutional module follows the specifications of \citet{kim2015character}. Table \ref{t:baseline} summarizes our baseline results. We use the BLEU metric \citep{papineni2002bleu} to compare our models. All scores are calculated on detokenized outputs using SacreBLEU\footnote{\url{https://github.com/mjpost/sacrebleu}} \citep{post-2018-call}.

As the table shows, no matter how powerful the engine is, adding even $5$\% noise is enough to break the model. Each segmentation scheme shows a unique behaviour. We were expecting a significant deterioration in the \textit{word2bpe} case but for both directions it provides relatively competitive results. \textit{bpe2bpe} seems to be the best as it gives the highest baseline where no noise is involved and shows less drop for noisy test sets. \textit{char2bpe} has the least decline when noise is added but it should be noted that it is not able to compete with others in the absence of noise. Although its degradation is minimal, it degrades from a non-optimal baseline. \textit{ConvT}, despite its sophisticated architecture, could not outperform \textit{bpe2bpe} and this was expected as tuning such a network over our (relatively) large dataset ({\fontfamily{pcr}\selectfont wmt-14} with $4.5$M samples) could be challenging.

\subsection{Results for Proposed Models}
In this section we report results for \textit{word2bpe}, \textit{bpe2bpe}, and \textit{char2bpe} settings when used with our solutions (\textit{TAFT}, \textit{CD}, and \textit{DCD}). 

\begin{table}[ht]
\centering
        \begin{tabular}{l c c c c c }
        \toprule
        \textbf{Model} & \textbf{0\%} & \textbf{5\%} & \textbf{10\%} & \textbf{20\%} & \textbf{30\%} \\\midrule
        \multicolumn{6}{c}{En$\rightarrow$De}\\
        \hline
        $\mathcal{T}^8_4$ & 28.48 & 22.21 & 17.05 & 10.28 & 5.99 \\ \hline
        \textit{FT} & 29.21 & 27.15 & 25.32 & \textbf{21.93} & \textbf{17.79} \\ 
        \textit{TAFT} & 29.47 & 27.33 & 25.35 & 21.33 & 17.29 \\ 
        \textit{CD} & 29.03 & 27.02 & 25.00 & 20.70 & 16.75 \\ 
        \textit{DCD} & \textbf{29.48} & \textbf{27.52} & \textbf{25.65} & 21.68 & 17.76 \\ \hline
        \multicolumn{6}{c}{De$\rightarrow$En}\\\hline
        $\mathcal{T}^8_4$  & 25.94 & 23.28 & 20.32 & 15.79 & 12.00 \\
        \hline
        \textit{FT} & \textbf{27.16} & \textbf{26.96} & 26.69  & 26.16 & 25.83 \\
        \textit{TAFT}  & 27.00 & 26.87 & \textbf{26.84} & \textbf{26.27} & \textbf{26.08} \\
        \textit{CD} & 27.1 & 26.83  & 26.61 & 26.04 & 25.57 \\
        \textit{DCD} & 27.06 & 26.86  & 26.83 & 26.13 & 26.03 \\
        \bottomrule
        \end{tabular} 
    \caption{\label{t:w2b} \textit{word2bpe} results. $\mathcal{T}^8_4$ is a Transformer with $8$ encoder and $4$ decoder layers. Boldfaced numbers are the best scores of each column.}
\end{table}

Table \ref{t:w2b} summarizes results related to \textit{word2bpe}. When translating into German, \textit{DCD} outperforms all other models where up to $10$\% noise is added to the test set. For extreme cases with $20$\% and $30$\% noise, \textit{FT} is more effective. In the opposite direction, \textit{FT} and our \textit{TAFT} extension provide the best performance. \textit{TAFT} also shows a very promising result for the $30$\% test set and even defeats the noise-free setting. The huge gap between the vanilla $\mathcal{T}^8_4$ and engines equipped  with our techniques shows the necessity of building noise-robust NMT models, specially if they are supposed to be deployed in real-world applications.

Results for \textit{bpe2bpe} models are reported in Table \ref{t:b2b}. For the En$\rightarrow$De direction, \textit{CD} is superior for all test sets, and this indicates that a loss function over a \textit{bpe}-based encoder could remarkably increase robustness. We observe a similar trend in the previous experiment for the De$\rightarrow$En direction. \textit{DCD} is also quite successful when test sets are fairly noisy. It seems \textit{bpe}-based Transformers benefit a lot from multi-tasking since both \textit{TAFT} and \textit{DCD} force the decoder to perform a second task in addition to translation.

\begin{table}[ht]
        \begin{tabular}{l c c c c c }
        \toprule
        \textbf{Model} & \textbf{0\%} & \textbf{5\%} & \textbf{10\%} & \textbf{20\%} & \textbf{30\%} \\\midrule
        \multicolumn{6}{c}{En$\rightarrow$De}\\
        \hline
        $\mathcal{T}^8_4$  & 28.46 & 24.82 & 21.58  & 15.98 & 11.89 \\
        \hline
        \textit{FT}  & 28.8 & 27.95 & 27.01 & 24.6 & 21.84 \\
        \textit{TAFT} & 28.96 & 28.03 & 26.65  & 24.02 & 21.16 \\
        \textit{CD}  & \textbf{29.49} & \textbf{28.51} & \textbf{27.68} & \textbf{25.27} & \textbf{22.63} \\
        \textit{DCD} & 28.91 & 28.02 & 26.89 & 24.37  & 21.47 \\ \hline
        \multicolumn{6}{c}{De$\rightarrow$En}\\\hline
        $\mathcal{T}^8_4$  & 28.04 & 24.87 & 21.61 & 16.11 & 11.48 \\
        \hline
        \textit{FT}   & 28.46 & 28.4 & 28.22  & 27.83 & 27.51 \\
        \textit{TAFT} & \textbf{28.73} & \textbf{28.53} & \textbf{28.51} & 27.93 & 27.63 \\
        \textit{CD}  & 28.52 & 28.42  & 28.25 & 27.84 & 27.50 \\
        \textit{DCD}  & 28.65 & 28.49  & 28.4 & \textbf{27.98} & \textbf{27.65}\\
        \bottomrule
        \end{tabular} 
        
    \caption{\label{t:b2b} \textit{bpe2bpe} results.}
\end{table}
\begin{table}[ht]
     \centering
        \begin{tabular}{l c c c c c }
        \toprule
        \textbf{Model} & \textbf{0\%} & \textbf{5\%} & \textbf{10\%} & \textbf{20\%} & \textbf{30\%} \\\midrule
        \multicolumn{6}{c}{En$\rightarrow$De}\\
        \hline
        $\mathcal{T}^8_4$ &  26.07 & 24.23 & 21.84 & 18.37 & 15.01 \\
        \textit{ConvT} & 25.46 & 22.55 & 20.13 & 14.9 & 11.29 \\
        \hline
        \textit{FT} & 27.24 & 26.50 & 25.92  & 24.51 & 23.36 \\
        \textit{TAFT}  & 27.11 & 26.41 & 25.56 & 23.90 & 22.14 \\
        \textit{CD}  & \textbf{27.29} & 26.5  & \textbf{26.05} & \textbf{24.71} & \textbf{23.37} \\
        \textit{DCD} & 27.2 & \textbf{26.73}  & 25.88 & 24.58 & 23.12 \\ \hline
        \multicolumn{6}{c}{De$\rightarrow$En}\\
        \hline
        $\mathcal{T}^8_4$ & 26.59 & 25.01 & 22.73 & 19.42 & 15.93 \\
        \textit{ConvT} & 27.08 & 24.01 & 21.39 & 16.44 & 11.59 \\
        \hline
        \textit{FT}  & 27.31 & 27.14 & 27.07  & 26.83 & 26.49 \\
        \textit{TAFT}  & 27.64 & \textbf{27.52} & 27.32 & 26.95 & 26.53 \\
        \textit{CD}  & 27.26 & 27.15  & 26.89 & 26.78 & 26.4 \\
        \textit{DCD}  & \textbf{27.71} & \textbf{27.52}  & \textbf{27.45} & \textbf{27.06} & \textbf{26.78}\\
        \bottomrule
    \end{tabular}
    \caption{\label{t:c2b} \textit{char2bpe} results.  \textit{ConvT} is added as additional baseline as encoders rely on characters.}
\end{table}
 
Finally, we summarize results of \textit{char2bpe} models in Table \ref{t:c2b}. Trends for this set of experiments are relatively consistent with previous ones. For the En$\rightarrow$De direction, the best result on average is delivered by \textit{CD} but \textit{DCD} also shows comparable performance. For the opposite direction, \textit{TAFT} and \textit{DCD} are better choices and multi-tasking again shows its impact. Because \textit{char2bpe} is a character-based model we also added results from \textit{ConvT} as another baseline to study if the convolutional operation can mitigate the problem and handle noise better. Experimental results demonstrate that there is no need for such a complex configuration and our techniques can train high-quality engines. 

\section{Conclusion and Future work}\label{cfw}
In this paper, we studied the problem of noise in the context of NMT and particularly focused on Transformers. We proposed three novel techniques to augment data and change the training procedure as well as the neural architecture. Experimental results show that our techniques can protect NMT engines from noise. Our models only affect the training phase and do not add any overhead in terms of space and/or time complexities at inference time. The findings of our research can be summarized as follows: 
\begin{itemize}
    \item There is no clear winner among our proposed models. Each approach has its own strength and should be adapted with respect to the problem. 
    
    \item \textit{FT} and \textit{TAFT} are data-driven techniques and can be applied to existing translation models with minimal effort.
    
    \item \textit{CD} and \textit{DCD} require some modifications in the neural architecture but they are able to provide promising results.
    
    \item Multi-tasking was quite useful in this scenario and it seems Transformers benefit a lot when their decoder is informed about source-side noise. 
\end{itemize}

In this research, we ran an extensive number of experiments in order to find the best configuration of each model and optimize hyper-parameters, but there still exist some unexplored topics/areas. In our future work, we are planning to experiment with other language pairs with different morphological and grammatical structures.
We are also interested in studying other noise classes. We could only afford to work with one class and we selected natural noise as we find it more realistic among others, but this work can be extended to other noise classes. Finally, our models are not unique to Transformer and NMT. We aim to evaluate them in other language processing/understanding tasks with other architectures.
\section*{Acknowledgements}
We would like to thank our anonymous reviewers as well as Mehdi Rezagholizadeh and Yimeng Wu for their valuable comments and feedback. 

\bibliography{emnlp2021}
\bibliographystyle{acl_natbib}




\end{document}